\documentclass{article}

    \PassOptionsToPackage{numbers, compress}{natbib}

\usepackage[final]{neurips_2022_ml4ps}

\usepackage[utf8]{inputenc} 
\usepackage[T1]{fontenc}    
\usepackage{hyperref}       
\usepackage{url}            
\usepackage{booktabs}       
\usepackage{amsfonts}       
\usepackage{nicefrac}       
\usepackage{microtype}      
\usepackage{xcolor}         
\usepackage{amsmath}
\usepackage{cleveref}

\usepackage{caption}
\usepackage{subcaption}
\usepackage{graphicx}
\graphicspath{ {./img/} }

\title{Graph Structure from Point Clouds: Geometric Attention is All You Need}

\author{
  Daniel Murnane\\
  Scientific Data Division\\
  Lawrence Berkeley National Laboratory\\
  Berkeley, CA USA \\
  \texttt{dtmurnane@lbl.gov} \\
  }

\begin{document}

\maketitle

\begin{abstract}
The use of graph neural networks has produced significant advances in point cloud problems, such as those found in high energy physics. The question of how to produce a graph structure in these problems is usually treated as a matter of heuristics, employing fully connected graphs or K-nearest neighbors. In this work, we elevate this question to utmost importance as the Topology Problem. We propose an attention mechanism that allows a graph to be constructed in a learned space that handles geometrically the flow of relevance, providing one solution to the Topology Problem. We test this architecture, called GravNetNorm, on the task of top jet tagging, and show that it is competitive in tagging accuracy, and uses far fewer computational resources than all other comparable models.
\end{abstract}

\section{Introduction}
\label{sec:intro}

Relational neural networks such as transformers and graph neural networks (GNNs) have pushed the limits of ML performance on many tasks, and the attention mechanism has been shown to be a key ingredient for achieving these state-of-the-art (SotA) results \cite{https://doi.org/10.48550/arxiv.1706.03762}. In natural language processing for example, attention-based transformers treat sentences as graphs, where words are represented by nodes and are "fully connected" (FC) - that is, all nodes are connected to all other nodes \cite{joshi2020transformers}. Attention-based GNNs have also been successfully employed in high energy physics (HEP) \cite{Shlomi_2021, https://doi.org/10.48550/arxiv.2012.08515, Qasim_2022, Pata_2021, Guo_2021, https://doi.org/10.48550/arxiv.2003.11603, ju2021performance, https://doi.org/10.48550/arxiv.2203.12852, hewes2021graph, iiyama2021distance}, a domain where data is often represented by point clouds of objects in space. The choice of how to connect point-cloud nodes into a graph is often non-obvious. The FC topology scales poorly with the complexity of the problem, possibly being prohibited by hardware constraints. Additionally, attention in many models is handled in a separate stage from the construction of the graph (the "choice of topology"), and is usually obtained as a learned function of pairs of node features (as in \cite{veličković2018graph}), which can be computationally expensive. In short, if not handled carefully, an attention mechanism applied to a fully-connected point cloud scales as $O(N_{nodes}^2)$.

In this work, we seek to address both of these hurdles - the choice of topology and the cost of attention - with a single solution. By adapting an existing architecture called GravNet \cite{Qasim_2019}, we propose an attention mechanism that is entirely dependent on a learned embedding space, and in doing so construct the topology of the graph in that space, at each iteration of message passing. The resulting network is called GravNetNorm as it extends GravNet to handle a subtle shortcoming of the original implementation, where the relevance of neighboring nodes was diffused through a mixture of geometry and node features. This required the use of a K-nearest-neighbor graph construction to function well. Our updated model instead learns the appropriate neighborhood size node-by-node, and in doing so uses fewer computational resources and performs with better accuracy than the original GravNet. Additionally, we apply GravNetNorm to a classic point cloud problem - jet flavor tagging - and show it is competitive with SotA methods, while taking an order of magnitude less memory, and a factor of four less time. We propose several extensions to this model that may improve accuracy further, while still retaining the learned geometric attention that makes it desirable for point cloud applications.

\section{Geometric Attention and the Topology Problem}
\label{sec:geo_att}

\subsection{Constructing a Graph}

Much work has been done in applying machine learning techniques to point cloud problems\cite{Qi_2017_CVPR}, and in particular attention models, typically for 3D points \cite{https://doi.org/10.48550/arxiv.2108.00620, https://doi.org/10.48550/arxiv.2102.10788, wang2019graph, https://doi.org/10.48550/arxiv.1909.12663}. We take as a case-study the problem of tagging jets of reconstructed particles as coming from either a top quark or a lighter hadronic particle \cite{toptaglandscape}. In this case as in most point cloud problems, we are given only a set of points (herein called "nodes"), each with a feature vector, but without any notion of inter-node connections or relationships (herein called "edges"). To apply a GNN to these problems, there are two limiting approaches. The first is to treat the nodes as unconnected - that is, as a set. The DeepSets architecture \cite{https://doi.org/10.48550/arxiv.1703.06114} has been used in jet tagging with, at the time, SotA results \cite{Hartman:2721094,Komiske_2019}. The other limit is to to treat the point cloud as fully connected, and this is the approach taken in transformer models, such as the Particle Transformer \cite{ParT}, which outperforms the set-limit approach in top tagging, although with significant computational overhead. A happy medium is struck by ParticleNet \cite{Qu_2020}, a model that applies a GNN to neighborhoods of K=16 neighbors and achieves very good results\footnote{Apples-to-apples comparisons are subtle, as training dataset size is a large factor in performance. See \cite{ParT} for a comprehensive analysis.}. Given these three working points (unconnected, fully-connected, and sparsely connected), we therefore suggest that including graph structure benefits a model's predictive power, but that most node-pair connections are not relevant to the prediction task. 

The attention mechanism addresses exactly this hypothesis. A multilayer perceptron (MLP), applied to pairs of nodes, learns which neighboring nodes carry relevant features and up-weight them in the message passing aggregation. The catch-22 is that nodes must be connected somehow in order to apply the weighted aggregation. The question of how to form edges we refer to as the Topology Problem: 

\begin{quote}
\itshape
    Given a variable sized set of nodes (a “point cloud”) and a loss function, then aside from a set of optima achieved by the learned GNN MLP weights, there is also a set of optima achieved by the topology of the attention based message passing.
\end{quote}

For a sparsely connected GNN, for a particular message passing step, it is non-obvious which nodes are most informational or relevant to other nodes. Many construction approaches, such as that used in ParticleNet, assume the best topology to be homophilic - that is, nodes with similar latent features should be topologically close. However, this is an arbitrary constraint, and some message steps may benefit from connections with dissimilar nodes\footnote{See \cite{https://doi.org/10.48550/arxiv.2202.07082} for a review of heterophily in graph neural networks.}. The solution is partly provided by having a second, independent latent space in which the graph is constructed. This is the mechanism adopted by GravNet and GarNet, two models proposed for GNN learning on point clouds. 

In the elegant approach suggested by the authors of GravNet, two latent spaces are learned for each node update step. The first is the hidden features to be aggregated, $h_i$. The second is an embedding space vector $\vec{s}_i \in S$ to be used to calculate the KNN neighborhood $K$ and attention weights $A_{ij}$. Both latent spaces are learned by MLPs applied independently to the input features of each GravNet convolution layer. The aggregated node features are thus given as 
\begin{align}
    h_i' = \sum_{j\in K} A(d_{ij}, h_j)\cdot \hat{h}_j , \qquad \text{where} \qquad A(d_{ij}, h_j) = |h_i|_{L1} e^{- G d_{ij}^2},  \qquad d_{ij} = |\vec{s}_i - \vec{s}_j|_{L2} \label{eq:gravconv}
\end{align}
where $G$ is a hyperparameter that acts like a gravitational constant. We define normalized hidden vectors $\hat{h}_i = h_i / |h_i|_{L1}$, using L1 norm.

\subsection{Geometry as Attention: GravNetNorm}

\begin{figure}
    \centering
    \begin{subfigure}[b]{0.49\textwidth}
        \centering
        \includegraphics[height=16em]{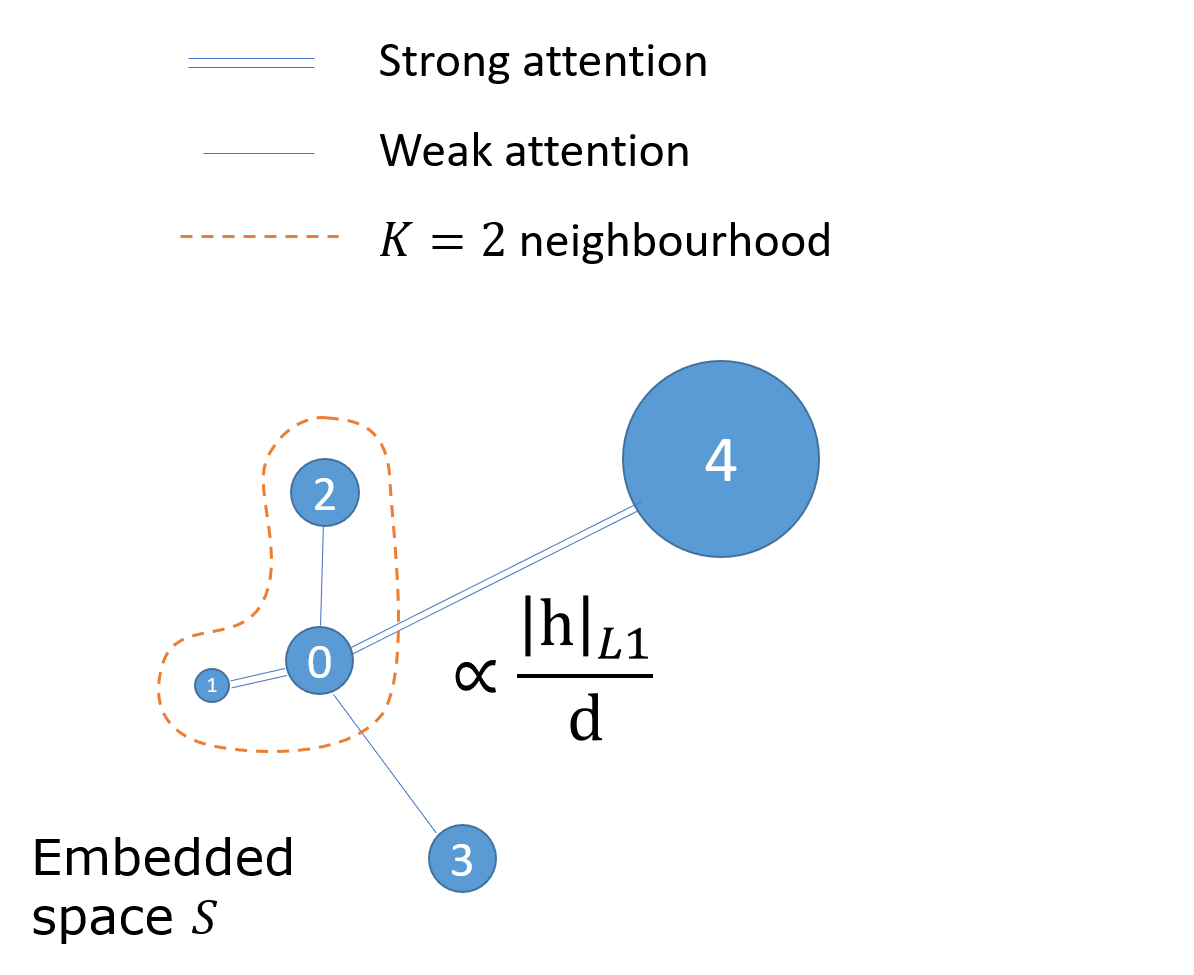}
        \caption{Original GravNet}
        \label{fig:massive_att}
    \end{subfigure}
    \begin{subfigure}[b]{0.49\textwidth}
        \centering
        \includegraphics[height=16em]{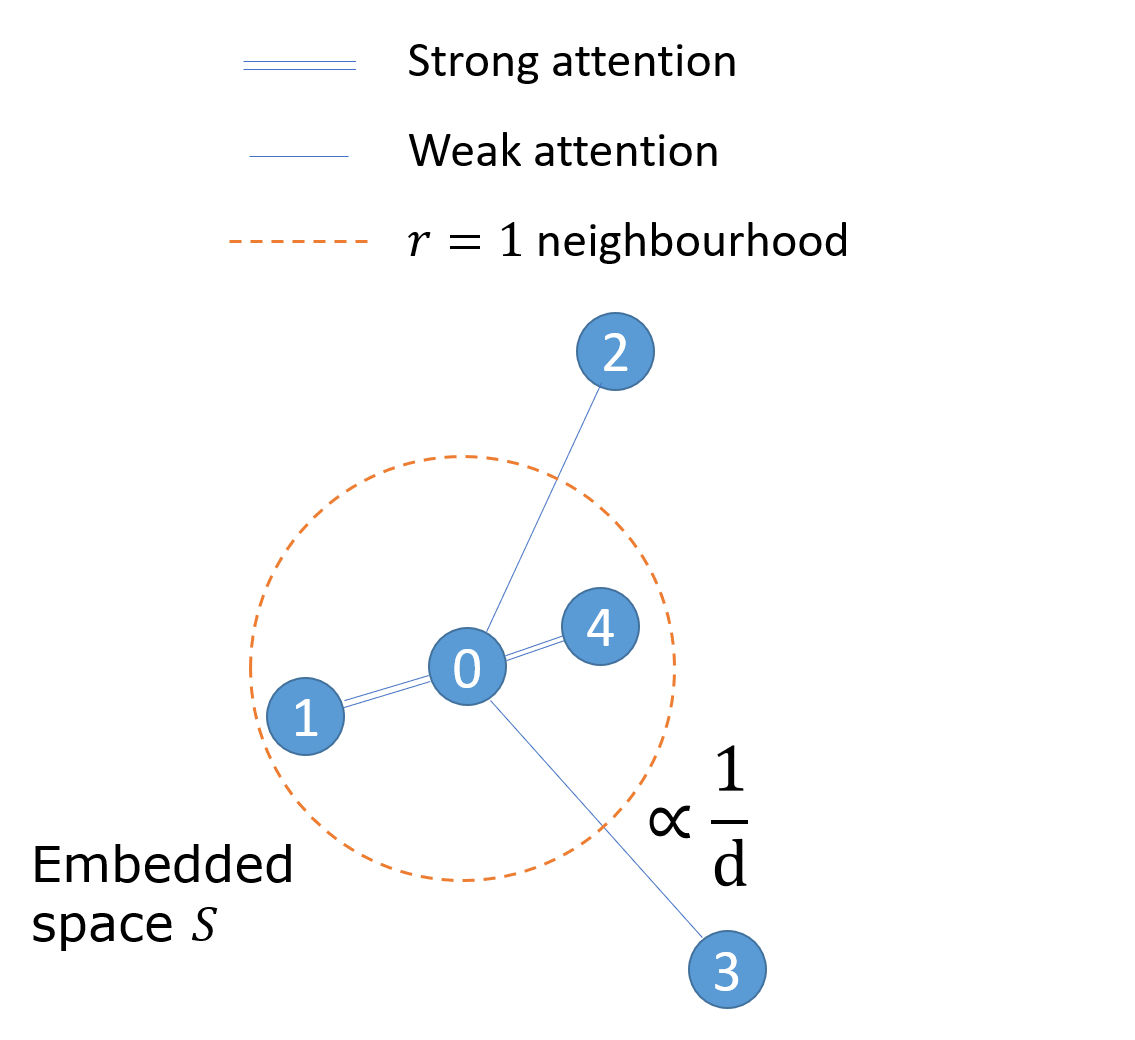}
        \caption{GravNetNorm}
        \label{fig:massless_att}
    \end{subfigure}
    \caption{Sketch of the GravNet attention mechanisms. The original GravNet node update propagates features $h$ proportionally to $|h|/d$, such that a node is affected by nearby (in embedded space $S$) \textit{and} heavy nodes. GravNetNorm constrains information to flow only through a function of distance, and therefore the geometry fully captures the attention mechanism. Thus only \textit{nearby} nodes need to be considered in the node update function.}
    \label{fig:geo_att}
\end{figure}

We are motivated to refine the GravNet architecture by the Topology Problem: Is the attention given to each neighboring node completely captured by the embedding space, and thus is an optimal topology constructed? Intuitively, we look at which information or relevance flows from one node to the next in message passing\footnote{One can formalize this intuition using Layerwise Relevance Propogation (LRP) analysis. An introduction to this is given in \cite{montavon2019layer} and an application to GNNs developed in \cite{https://doi.org/10.48550/arxiv.2111.12840}. The full calculation of LRP in geometry-constrained attention will be provided in an upcoming study.}. In the original GravNet model, nodes are influenced proportionally to both the closeness of a neighbor $d_{ij}$ and the \textit{size} of a neighbor $|h_j|_{L1}$. This is sketched in \cref{fig:massive_att}. The latter $|h_j|_{L1}$ factor means that a distant neighbor may still have an oversized influence if it is an "important" node (whatever this may mean in the problem being considered). Thus, a graph constructed according to nearness in $S$ will not necessarily reflect the attention function, leading to important connections possibly being missed, and a suboptimal solution to the Topology Problem. Flow of information as a function of both neighbor size and distance is well-defined in a FC graph, hence the excellent performance of transformers. However if we require a sparse topology, we need to know which neighbors to connect. In the GravNet case, they will be connections that maximize $\frac{\textnormal{size}}{\textnormal{distance}}$ - an expensive calculation needing to be made across all pairs.  Instead, if the weighting of information is only a function of distance, we only need to consider neighbors within a radius $r$ in $S$, which can be calculated efficiently and scales well with graph size.

The solution is simple: Normalize hidden features such that all nodes have a total size of 1, and therefore constrain the GNN to pass all relevance through the geometry of $S$ alone. That is, we take 
\begin{align}
A(d_{ij}, h_j) = \exp(- G \frac{d_{ij}^2}{r^2})    
\end{align}

Although a seemingly minor alteration, this produces a most-minimal implementation of geometry-constrained attention mechanism. We also introduce a factor $\frac{1}{r^2}$ in the attention function. This new hyperparameter $r$ appears in the following training procedure: Assuming now that all attention is constrained to the neighbourhood of each node in $S$, we should train and inference our model using topology built from that neighbourhood only. That is, we construct a radius graph in each message passing step, with radius $r$. Once this $r$ hyperparameter is set, e.g. to $r=1$, then the gravitational constant $G$ can then be used to tune the sparsity of the topology. E.g. a choice of $G=3$ means that nodes at distance $r=1$ will be given an attention weight of around $0.05$. For the problem considered here, this seems to be the choice of G above which performance plateaus. The effect of normalizing node sizes is sketched in \cref{fig:massless_att}. Note that the embedding space $S$ need not be normalized, so we continue to use Euclidean distance as the learned attention function. The details of the implementation and the training procedure are available in a public Github repository \cite{MurnaneGithub}.



\section{Results}

\subsection{Top Tagging Problem}

The dataset used in this study is made available in \cite{kasieczka2019top}, which contains a set of 1.4m training jets, and 400k each of validation and test jet samples. A jet contains up to 200 constituent reconstructed particle 4-vectors, which we take as nodes. A further 17 hand-engineered features are attached to each node, taken to match those described in \cite{Qu_2020}. The task of top tagging is to classify each jet as either originating from the decay of a top quark, or from the decay of a lighter quark or gluon. We thus treat this as a graph-level binary classification problem, where the GNN must output a classification score between 0 and 1 for each graph, which is used in a binary cross entropy loss function, with no positive weighting as the dataset is well-balanced.

\subsection{Physics Performance}

An initial study of the physics performance of the original GravNet and GravNetNorm is presented in \cref{tab:phys_perf}, along with several other high-performing deep neural networks\footnote{A note on ParticleNet performance: This is the published performance. We were not able to obtain this result. The training techniques used in that work could also be used to improve GravNetNorm performance.}. Both the accuracy and area under the ROC curve (AUC) are given, as well as the background rejection rate $\epsilon_B^{-1}$ (where $\epsilon_B$ is the false positive rate) at a working point of $30\%$ efficiency. 

One can see that GravNetNorm outperforms all other models, except for ParticleNet. This shortcoming in performance can be attributed to several factors. The first is that layer sizes are heuristically taken from existing models, and may not be optimally suited to this new architecture. Additionally, in training, we note significant overfitting even on the full training set of 1.2 million jets and with a dropout of 0.2. Performance plateaus above this dropout rate. As such, we propose in an upcoming work to use a larger dataset such as that created in \cite{ParT}, to fully explore the predictive power of GravNetNorm. One can also see in the table that the original GravNet performs well, but not equivalently with the updated variant.

Further improvements are being studied, and will be presented in a near-future work, to boost the physics performance of GravNetNorm. These include dividing the spatial vector to use as a multi-headed attention (a mechanism implicit in the ParticleNet architecture), and learning dynamically the \textit{number} of message passing steps each node requires, just as we do with the number of topological neighbors. These will both add expressiveness without losing the geometry-constrained attention mechanism.

\begin{table}
 
  \centering
  \begin{tabular}{llll}
    \toprule
    Model  & Acc & AUC & $\epsilon^{-1}_B|_{30\%}$ \\
    \midrule   
    P-CNN & 0.936 & 0.9837 & 1174 $\pm$ 58 \\
    PFN & 0.932 & 0.9819 & 888 $\pm$ 17 \\
    Gravnet & 0.937 & 0.9844 & 1340 $\pm$ 69 \\
    ParticleNet & 0.940 & 0.9858 & 1615 $\pm$ 93 \\
    \midrule 
    GravnetNorm & 0.939 & 0.9850 & 1438 $\pm$ 35 \\
    \bottomrule
  \end{tabular}
  \vspace{1em}
   \caption{Comparison of top tagging physics performance for a selection of DNNs \cite{toptaglandscape, CMS-DP-2017-049, Komiske_2019}. The performance of the first three models is quoted from \cite{Qu_2020}, and all results are averaged across five training runs. Variation across these runs is given for background rejection, while variation of accuracy and AUC is negligible. Other high-performing taggers (\cite{Gong_2022,ParT}) are not compared here as they contain features orthogonal to geometric attention, such as equivariance. Future work will seek to combine these mechanisms.}
  \label{tab:phys_perf}
\end{table}

\subsection{Computational Performance}
\label{sec:comp_perf}

\begin{table}
 
  \centering
  \begin{tabular}{llll}
    \toprule
    Model     & \# Parameters & Max. memory (Gb) & Time ($\mu s$ per jet) \\
    \midrule   
    P-CNN & 348k & - & 110 \\
    PFN & 82k & - & 120 \\
    ParticleNet & 467k &  3.1 & 88 \\
    Gravnet & 545k & 0.87   & 37  \\
    \midrule
    GravnetNorm & 545k & \textbf{0.23} & \textbf{22} \\
    \bottomrule
    
  \end{tabular}
  \vspace{1em}
   \caption{Comparison of memory and time requirements of top taggers. Best performances are given in bold. Performance is measured on an Nvidia 40Gb A100, with batch size 1000. Timings are given \textit{per jet}, that is $t_{jet} = t_{batch} / 1000$. The first two model timings are quoted from \cite{Gong_2022}.}
  \label{tab:comp_perf}
\end{table}

Inference performance is here measured by both the peak memory usage (taken as a proxy for the kind of hardware limitation these models may impose), and the average jet inference time in microseconds. Presented in \cref{tab:comp_perf}, we see that GravNetNorm is by far the most computationally efficient. Despite having a comparable number of parameters to other DNNs, this model has two features that allow superior performance. The first is the geometric attention mechanism. Since attention is learned node-wise in embedded space, the embedding step (i.e. the forward pass from $h_i \rightarrow \vec{s}_i$) scales as $O(N_{nodes})$. We see that both GravNet variants benefit from this. Compare this with the standard edge-wise attention, such as that employed in ParticleNet, which scales as $O(N_{edges})$. 

The second feature is that the topology is completely learned, so neighborhoods are only as large as required for good performance\footnote{It is indeed the case that the attention varies smoothly with the geometry, so some arbitrary choice of radius still needs to be made. However, we can quantify exactly the relevance of nodes outside this radius by $e^{-G}$, which is less than 5\% for $G=3$}. This allows GravNetNorm to consume fewer resources than GravNet. In particular, a radius graph construction scales naively as $O(k N_{nodes})$ (where k is the average neighborhood size), while a KNN construction requires neighbors to be sorted and scales naively as $O(N_{nodes}^2)$\cite{zhang2013fast}. The particular implementation used here is from Pytorch Cluster \cite{torch_cluster}, but performance can be boosted further for large point clouds with dedicated radius-graph algorithms \cite{Lazar_2023}. 

Additionally, K values are set arbitrarily by hand, but GravNetNorm learns to build neighborhoods of mean size [3, 8, 13] (in the top tagging case, in order of node update step), significantly improving the throughput of both the graph-building and aggregation operations. While hyperparameter tuning of K may improve a KNN-based model throughput - as there appear to be optimal choices of neighborhood size - this would still be a static value, rather than dynamic from point-to-point and event-to-event.

\section{Conclusion}
\label{sec:conclusion}

In this work, we have explored a long-standing obstacle in the application of graph neural networks to point clouds, which we term the Topology Problem. We present one set of solutions to this, in the form of a geometry-constrained attention. In particular, we alter the pre-existing GravNet architecture to construct a minimal geometric attention model, and show how it intuitively leads to a topology that captures the node connections with highest attention. We have taken an example use case to be graph-level top tagging, however the use of geometric attention could be applied to node-level or edge-level prediction tasks, and we will present results on those tasks in upcoming work. We show that our GravNetNorm variation is competitive in tagging accuracy with other state-of-the-art taggers, while requiring far fewer computational resources. As this is the "most-minimal" geometric attention model, future work will present techniques to combine geometric attention with other SotA architectures to further boost tagging accuracy. The codebase is available on Github \cite{MurnaneGithub}.

\newpage

\section{Impact Statement}
\label{sec:impact}

In this work, we propose several ideas that we hope will stimulate further discussion and research directions. These include:
\begin{itemize}
    \item A presentation of the \textbf{Topology Problem} - an oft-overlooked issue that is usually solved ad hoc in graph neural network application to point clouds. In reality, as high energy physics datasets grow in size and complexity, a careful analysis of how graph topology is constructed will be essential to scaling up production-ready models in collider and astroparticle experiments.
    \item A \textbf{geometry-constrained attention operator}, as applied in an amended version of the GravNet architecture. This can be seen as a most-minimal construction of a GNN that propagates all relevance entirely through geometry, and may open the door to more sophisticated attention geometries. Regardless, the operation as presented here can be dropped into existing architectures to greatly improve computational efficiency.
    \item Some suggestions are given of further exploration of geometric-constrained attention, including multi-headed attention and learned number of message passing iterations.
\end{itemize}

We do not expect this work to have any negative societal or ethical impacts.

\section{Acknowledgements}
\label{sec:ack}

This work is supported by the US DoE’s Office of Science, under contract \# DE-AC02-05CH11231 (CompHEP Exa.TrkX) and the Exascale Computing Project (17-SC-20-SC). This research used resources of the National Energy Research Scientific Computing Center (NERSC), a U.S. Department of Energy Office of Science User Facility located at Lawrence Berkeley National Laboratory, operated under Contract No. DE-AC02-05CH11231. I am grateful to Paolo Calafiura for comments on this work, as well as Ryan Liu, Gage DeZoort and Tuan Pham for discussions.

\bibliographystyle{unsrt}
\bibliography{refs}

\begin{thebibliography}{10}

\bibitem{https://doi.org/10.48550/arxiv.1706.03762}
Ashish Vaswani, Noam Shazeer, Niki Parmar, Jakob Uszkoreit, Llion Jones,
  Aidan~N. Gomez, Lukasz Kaiser, and Illia Polosukhin.
\newblock Attention is all you need, 2017.

\bibitem{joshi2020transformers}
Chaitanya Joshi.
\newblock Transformers are graph neural networks.
\newblock {\em The Gradient}, page~5, 2020.

\bibitem{Shlomi_2021}
Jonathan Shlomi, Peter Battaglia, and Jean-Roch Vlimant.
\newblock Graph neural networks in particle physics.
\newblock {\em Machine Learning: Science and Technology}, 2(2):021001, jan
  2021.

\bibitem{https://doi.org/10.48550/arxiv.2012.08515}
Yogesh Verma and Satyajit Jena.
\newblock Particle track reconstruction using geometric deep learning, 2020.

\bibitem{Qasim_2022}
Shah~Rukh Qasim, Nadezda Chernyavskaya, Jan Kieseler, Kenneth Long, Oleksandr
  Viazlo, Maurizio Pierini, and Raheel Nawaz.
\newblock End-to-end multi-particle reconstruction in high occupancy imaging
  calorimeters with graph neural networks.
\newblock {\em The European Physical Journal C}, 82(8), aug 2022.

\bibitem{Pata_2021}
Joosep Pata, Javier Duarte, Jean-Roch Vlimant, Maurizio Pierini, and Maria
  Spiropulu.
\newblock {MLPF}: efficient machine-learned particle-flow reconstruction using
  graph neural networks.
\newblock {\em The European Physical Journal C}, 81(5), may 2021.

\bibitem{Guo_2021}
Jun Guo, Jinmian Li, Tianjun Li, and Rao Zhang.
\newblock Boosted higgs boson jet reconstruction via a graph neural network.
\newblock {\em Physical Review D}, 103(11), jun 2021.

\bibitem{https://doi.org/10.48550/arxiv.2003.11603}
Xiangyang Ju, Steven Farrell, Paolo Calafiura, Daniel Murnane, Prabhat, Lindsey
  Gray, Thomas Klijnsma, Kevin Pedro, Giuseppe Cerati, Jim Kowalkowski, Gabriel
  Perdue, Panagiotis Spentzouris, Nhan Tran, Jean-Roch Vlimant, Alexander
  Zlokapa, Joosep Pata, Maria Spiropulu, Sitong An, Adam Aurisano, Jeremy
  Hewes, Aristeidis Tsaris, Kazuhiro Terao, and Tracy Usher.
\newblock Graph neural networks for particle reconstruction in high energy
  physics detectors, 2020.

\bibitem{ju2021performance}
Xiangyang Ju, Daniel Murnane, Paolo Calafiura, Nicholas Choma, Sean Conlon,
  Steven Farrell, Yaoyuan Xu, Maria Spiropulu, Jean-Roch Vlimant, Adam
  Aurisano, et~al.
\newblock Performance of a geometric deep learning pipeline for hl-lhc particle
  tracking.
\newblock {\em The European Physical Journal C}, 81(10):1--14, 2021.

\bibitem{https://doi.org/10.48550/arxiv.2203.12852}
Savannah Thais, Paolo Calafiura, Grigorios Chachamis, Gage DeZoort, Javier
  Duarte, Sanmay Ganguly, Michael Kagan, Daniel Murnane, Mark~S. Neubauer, and
  Kazuhiro Terao.
\newblock Graph neural networks in particle physics: Implementations,
  innovations, and challenges, 2022.

\bibitem{hewes2021graph}
Jeremy Hewes, Adam Aurisano, Giuseppe Cerati, Jim Kowalkowski, Claire Lee,
  Wei-keng Liao, Alexandra Day, Ankit Agrawal, Maria Spiropulu, Jean-Roch
  Vlimant, et~al.
\newblock Graph neural network for object reconstruction in liquid argon time
  projection chambers.
\newblock In {\em EPJ Web of Conferences}, volume 251, page 03054. EDP
  Sciences, 2021.

\bibitem{iiyama2021distance}
Yutaro Iiyama, Gianluca Cerminara, Abhijay Gupta, Jan Kieseler, Vladimir
  Loncar, Maurizio Pierini, Shah~Rukh Qasim, Marcel Rieger, Sioni Summers,
  Gerrit Van~Onsem, et~al.
\newblock Distance-weighted graph neural networks on fpgas for real-time
  particle reconstruction in high energy physics.
\newblock {\em Frontiers in big Data}, 3:598927, 2021.

\bibitem{veličković2018graph}
Petar Veličković, Guillem Cucurull, Arantxa Casanova, Adriana Romero, Pietro
  Liò, and Yoshua Bengio.
\newblock Graph attention networks, 2018.

\bibitem{Qasim_2019}
Shah~Rukh Qasim, Jan Kieseler, Yutaro Iiyama, and Maurizio Pierini.
\newblock Learning representations of irregular particle-detector geometry with
  distance-weighted graph networks.
\newblock {\em The European Physical Journal C}, 79(7), jul 2019.

\bibitem{Qi_2017_CVPR}
Charles~R. Qi, Hao Su, Kaichun Mo, and Leonidas~J. Guibas.
\newblock Pointnet: Deep learning on point sets for 3d classification and
  segmentation.
\newblock In {\em Proceedings of the IEEE Conference on Computer Vision and
  Pattern Recognition (CVPR)}, July 2017.

\bibitem{https://doi.org/10.48550/arxiv.2108.00620}
Shi Qiu, Yunfan Wu, Saeed Anwar, and Chongyi Li.
\newblock Investigating attention mechanism in 3d point cloud object detection,
  2021.

\bibitem{https://doi.org/10.48550/arxiv.2102.10788}
Xu~Wang, Yi~Jin, Yigang Cen, Tao Wang, and Yidong Li.
\newblock Attention models for point clouds in deep learning: A survey, 2021.

\bibitem{wang2019graph}
Lei Wang, Yuchun Huang, Yaolin Hou, Shenman Zhang, and Jie Shan.
\newblock Graph attention convolution for point cloud semantic segmentation.
\newblock In {\em Proceedings of the IEEE/CVF conference on computer vision and
  pattern recognition}, pages 10296--10305, 2019.

\bibitem{https://doi.org/10.48550/arxiv.1909.12663}
Mingtao Feng, Liang Zhang, Xuefei Lin, Syed~Zulqarnain Gilani, and Ajmal Mian.
\newblock Point attention network for semantic segmentation of 3d point clouds,
  2019.

\bibitem{toptaglandscape}
Gregor Kasieczka, Tilman Plehn, Anja Butter, Kyle Cranmer, Dipsikha Debnath,
  Barry Dillon, Malcolm Fairbairn, Darius Faroughy, Wojtek Fedorko, Christophe
  Gay, Loukas Gouskos, Jernej Kamenik, Patrick Komiske, Simon Leiss, Alison
  Lister, Sebastian Macaluso, Eric Metodiev, Liam Moore, Benjamin Nachman, and
  Sreedevi Varma.
\newblock The machine learning landscape of top taggers.
\newblock {\em SciPost Physics}, 7, 07 2019.

\bibitem{https://doi.org/10.48550/arxiv.1703.06114}
Manzil Zaheer, Satwik Kottur, Siamak Ravanbakhsh, Barnabas Poczos, Ruslan
  Salakhutdinov, and Alexander Smola.
\newblock Deep sets, 2017.

\bibitem{Hartman:2721094}
Nicole~Michelle Hartman, Michael Kagan, and Rafael Teixeira De~Lima.
\newblock {Deep Sets for Flavor Tagging on the ATLAS Experiment}.
\newblock Technical report, CERN, Geneva, 2020.

\bibitem{Komiske_2019}
Patrick~T. Komiske, Eric~M. Metodiev, and Jesse Thaler.
\newblock Energy flow networks: deep sets for particle jets.
\newblock {\em Journal of High Energy Physics}, 2019(1), jan 2019.

\bibitem{ParT}
Huilin Qu, Congqiao Li, and Sitian Qian.
\newblock Particle transformer for jet tagging, 2022.

\bibitem{Qu_2020}
Huilin Qu and Loukas Gouskos.
\newblock Jet tagging via particle clouds.
\newblock {\em Physical Review D}, 101(5), mar 2020.

\bibitem{https://doi.org/10.48550/arxiv.2202.07082}
Xin Zheng, Yixin Liu, Shirui Pan, Miao Zhang, Di~Jin, and Philip~S. Yu.
\newblock Graph neural networks for graphs with heterophily: A survey, 2022.

\bibitem{montavon2019layer}
Gr{\'e}goire Montavon, Alexander Binder, Sebastian Lapuschkin, Wojciech Samek,
  and Klaus-Robert M{\"u}ller.
\newblock Layer-wise relevance propagation: an overview.
\newblock {\em Explainable AI: interpreting, explaining and visualizing deep
  learning}, pages 193--209, 2019.

\bibitem{https://doi.org/10.48550/arxiv.2111.12840}
Farouk Mokhtar, Raghav Kansal, Daniel Diaz, Javier Duarte, Joosep Pata,
  Maurizio Pierini, and Jean-Roch Vlimant.
\newblock Explaining machine-learned particle-flow reconstruction, 2021.

\bibitem{MurnaneGithub}
Daniel Murnane.
\newblock Geometric attention.
\newblock \url{https://github.com/murnanedaniel/GeometricAttention}, 2022.

\bibitem{kasieczka2019top}
Gregor Kasieczka, Tilman Plehn, Jennifer Thompson, and Michael Russel.
\newblock Top quark tagging reference dataset.
\newblock {\em Version v0 (2018\_03\_27). Mar}, 2019.

\bibitem{CMS-DP-2017-049}
{Boosted jet identification using particle candidates and deep neural
  networks}.
\newblock 2017.

\bibitem{Gong_2022}
Shiqi Gong, Qi~Meng, Jue Zhang, Huilin Qu, Congqiao Li, Sitian Qian, Weitao Du,
  Zhi-Ming Ma, and Tie-Yan Liu.
\newblock An efficient lorentz equivariant graph neural network for jet
  tagging.
\newblock {\em Journal of High Energy Physics}, 2022(7), jul 2022.

\bibitem{zhang2013fast}
Yan-Ming Zhang, Kaizhu Huang, Guanggang Geng, and Cheng-Lin Liu.
\newblock Fast knn graph construction with locality sensitive hashing.
\newblock In {\em Joint European Conference on Machine Learning and Knowledge
  Discovery in Databases}, pages 660--674. Springer, 2013.

\bibitem{torch_cluster}
Matthias Fey.
\newblock Pytorch cluster, 2023.

\bibitem{Lazar_2023}
Alina Lazar, Xiangyang Ju, Daniel Murnane, Paolo Calafiura, Steven Farrell,
  Yaoyuan Xu, Maria Spiropulu, Jean-Roch Vlimant, Giuseppe Cerati, Lindsey
  Gray, Thomas Klijnsma, Jim Kowalkowski, Markus Atkinson, Mark Neubauer, Gage
  DeZoort, Savannah Thais, Shih-Chieh Hsu, Adam Aurisano, Jeremy Hewes,
  Alexandra Ballow, Nirajan Acharya, Chun yi~Wang, Emma Liu, and Alberto Lucas.
\newblock Accelerating the inference of the exa.trkx pipeline.
\newblock {\em Journal of Physics: Conference Series}, 2438(1):012008, feb
  2023.

\end{thebibliography}

\end{document}